\documentclass[10pt,letterpaper]{article}
\usepackage[top=0.85in,left=2.75in,footskip=0.75in]{geometry}

\usepackage{amsmath,amssymb}

\usepackage{changepage}

\usepackage[utf8x]{inputenc}

\usepackage{textcomp,marvosym}

\usepackage{cite}

\usepackage{nameref,hyperref}

\usepackage[right]{lineno}

\usepackage{microtype}
\DisableLigatures[f]{encoding = *, family = * }

\usepackage[table]{xcolor}

\usepackage{array}

\newcolumntype{+}{!{\vrule width 2pt}}

\newlength\savedwidth

\raggedright
\usepackage[aboveskip=1pt,labelfont=bf,labelsep=period,justification=raggedright,singlelinecheck=off]{caption}

\makeatletter
\renewcommand{\@biblabel}[1]{\quad#1.}
\makeatother

\usepackage{lastpage,fancyhdr,graphicx}
\usepackage{epstopdf}
\fancyheadoffset[L]{2.25in}
\fancyfootoffset[L]{2.25in}
\usepackage{booktabs}
\usepackage{subcaption}
\usepackage{comment}
\usepackage{enumitem}

\begin{document}
\vspace*{0.2in}

% Title must be 250 characters or less.
\begin{flushleft}
{\Large
\textbf\newline{A new approach to evaluating legibility: Comparing legibility frameworks using framework-independent robot motion trajectories} 
% Please use "sentence case" for title and headings (capitalize only the first word in a title (or heading), the first word in a subtitle (or subheading), and any proper nouns).
}
\newline
% Insert author names, affiliations and corresponding author email (do not include titles, positions, or degrees).
\\
Wallkotter Sebastian\textsuperscript{*,1},
Mohamed Chetouani\textsuperscript{2},
Ginevra Castellano\textsuperscript{1},
\\
\bigskip
\textbf{1} Department of Information Technology, Uppsala University, Uppsala, Sweden
\\
\textbf{2} Institute for Intelligent Systems and Robotics, Sorbonne University, CNRS UMR 7222, Paris, France
\\
\bigskip

% Use the asterisk to denote corresponding authorship and provide email address in note below.
* sebastian.wallkotter@it.uu.se

\end{flushleft}
% Please keep the abstract below 300 words
\section*{Abstract}
Robots that share an environment with humans may communicate their intent using a variety of different channels. Movement is one of these channels and, particularly in manipulation tasks, intent communication via movement is called legibility. It alters a robot's trajectory to make it intent expressive. Here we propose a novel evaluation method that improves the data efficiency of collected experimental data when benchmarking approaches generating such legible behavior. The primary novelty of the proposed method is that it uses trajectories that were generated independently of the framework being tested. This makes evaluation easier, enables N-way comparisons between approaches, and allows easier comparison across papers. We demonstrate the efficiency of the new evaluation method by comparing $10$ legibility frameworks in $2$ scenarios. The paper, thus, provides readers with (1) a novel approach to investigate and/or benchmark legibility, (2) an overview of existing frameworks, (3) an evaluation of $10$ legibility frameworks (from $6$ papers), and (4) evidence that viewing angle and trajectory progression matter when users evaluate the legibility of a motion.

% \linenumbers

% For figure citations, please use "Fig" instead of "Figure".
% Place figure captions after the first paragraph in which they are cited.
% Place tables after the first paragraph in which they are cited.
%PLOS does not support heading levels beyond the 3rd (no 4th level headings).
\section{Introduction}\label{sec:introduction}
Legible robot motion is motion that, as the name suggests, is clear to read, i.e., observers can easily understand the robot's intentions by observing its movement. The opposite, illegible robot motion, is a movement that (willing or incidental) obfuscates a robot's intentions by making it hard for an observer to read the movement. Most movements fall between these two extremes; they are somewhat legible, but it is also possible to find more legible (or illegible) alternatives. In other words, there exists a spectrum of legible movement.

Robots that can navigate this spectrum of legible movement have several advantages over their less legible peers. Increasing legibility increases the robot's transparency, which often has positive effects on measures of trust, robustness, and efficiency \cite{wallkotter2020}. As a form of non-verbal communication, legibility may also increase the quality of human-robot interaction \cite{dautenhahn2005}, e.g., by establishing common ground \cite{cha2018}.

Assuming these advantages seem worthwhile, we would like to know how to create such legible motion. For this, previous work proposes several frameworks of which some have been reviewed previously, either in a dedicated review on legibility \cite{lichtenhaler2016}, in an effort to generalize legibility \cite{sreedharan2021}, or in one of several reviews of related areas \cite{chakraborti2018b,cha2018,venture2019}. 

While such reviews provide an overview of existing methods, they don't specify how to choose between existing frameworks. Yet this choice can become important, for example if we wish to study the impact of legibility. Here, choosing the most legible framework for a scenario increases the observable effect and hence reduces the burden on sample size.

This is the central topic of this paper: How should we decide which framework to use? I.e., how can we benchmark different legibility frameworks? To answer this question we propose a new evaluation method. Instead of relying on pairwise comparisons of framework-optimal trajectories (traditional method, see sec. \ref{sec:traditional-approach}), the method proposed here performs a direct N-way comparison of frameworks by relying on a fixed set of framework-independent trajectories. Using this new method allows us to rank the quality of existing legibility methods for specific scenarios without having to perform extensive pair-wise experiments.

\begin{figure}[t]
    \centering
    \includegraphics[width=0.5\textwidth]{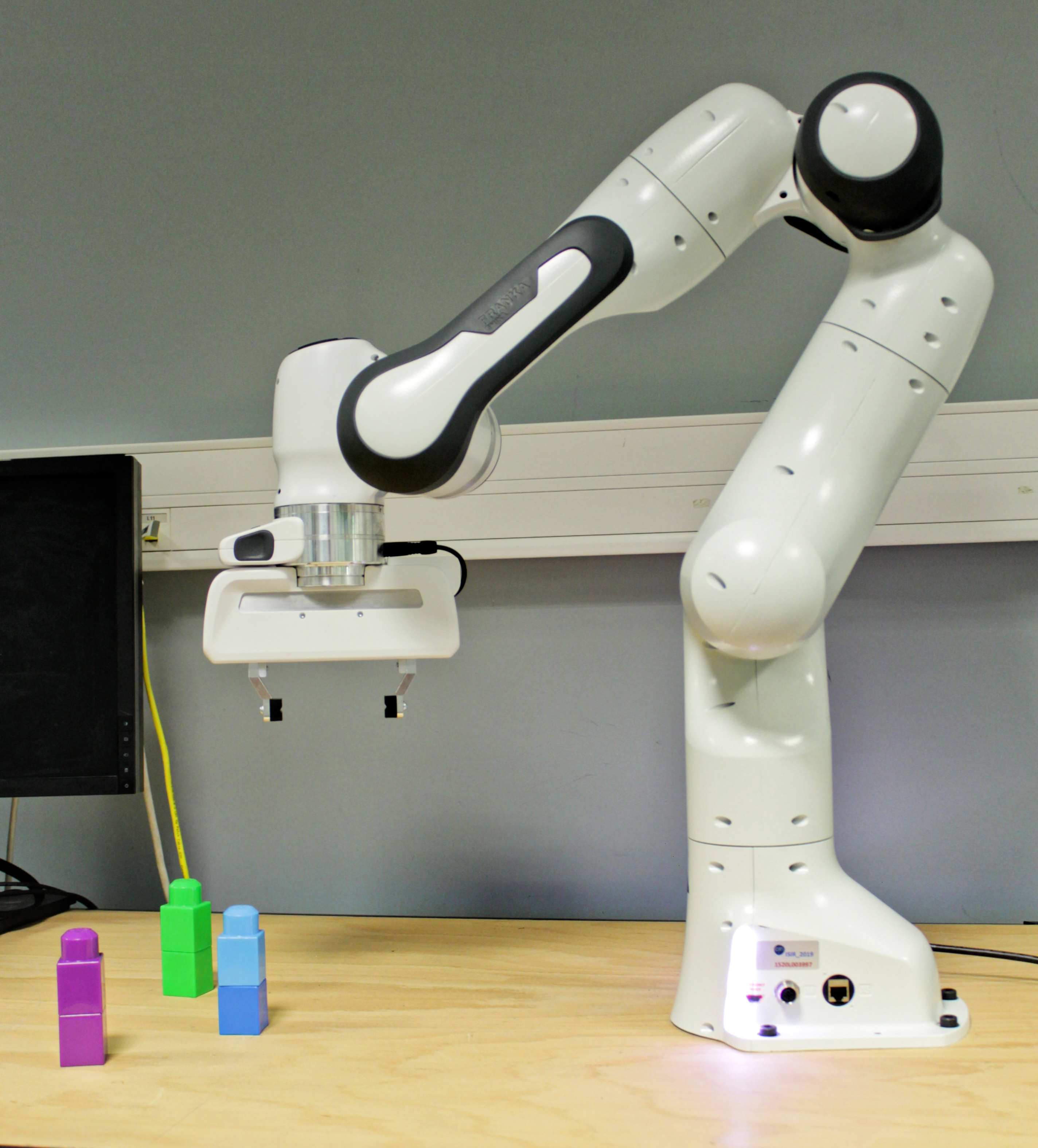}
    \caption{{\bf Video Dataset Experimental Setup (Scenario 1)} The picture shows a Franka Emika that is tasked with picking up one of three colored objects (violet, blue, green) while being legible to an observer that views the motion from the angle shown in the picture.}
    \label{fig:video-data-teaser}
\end{figure}

\begin{figure}[t]
    \centering
    \includegraphics[width=\textwidth]{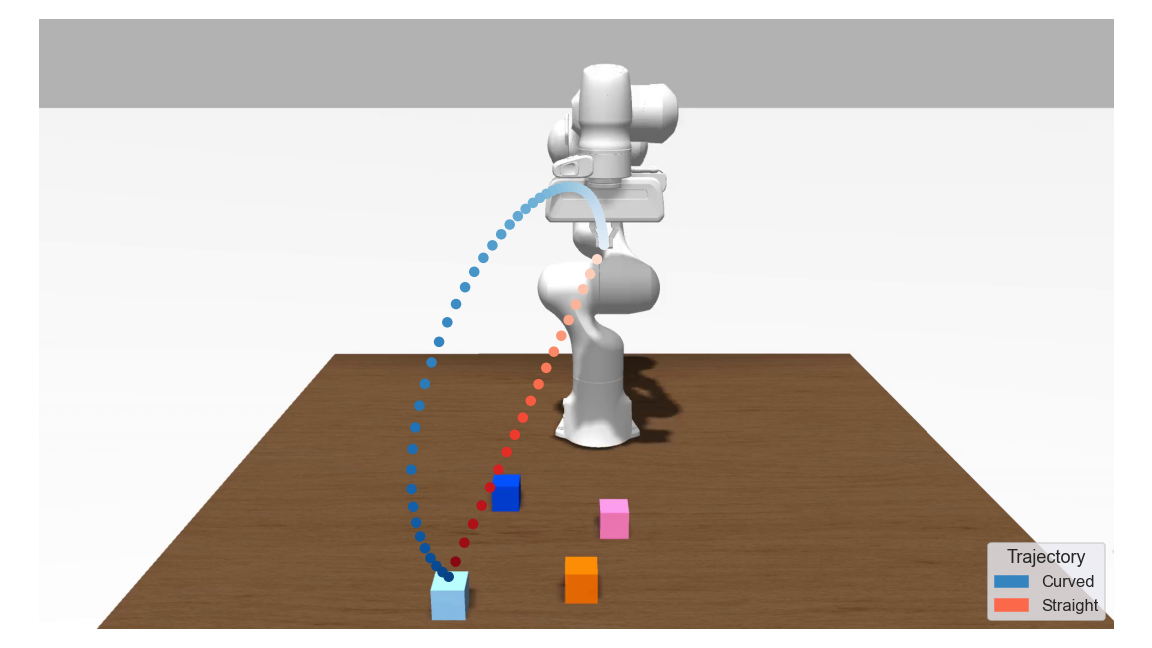}
    \caption{{\bf Simulation Dataset Experimental Setup (Scenario 2)} The picture shows a simulated Franka Emika robot that is tasked with picking up one of four colored cubes (light blue, dark blue, yellow, pink) while being legible to an observer that views the motion from the angle shown in the picture. In this scenario, there are two additional angles, which are shown in the appendix.}
    \label{fig:simulation-data-teaser}
\end{figure}

To demonstrate the effectiveness of this method we collect data in a manipulation task (pick-up) for two slightly different scenarios. In the first scenario, we are interested in making a robot move legibly with respect to an observer looking at the scene from a single fixed angle. For this, we record videos of a physical robot. In the second scenario, we are interested in making a robot move legibly with respect to an observer looking at the scene from one of three different angles. For this, we generated videos of the behavior using a robot simulator. In both cases, we use a Franka Emika\footnote{https://www.franka.de/robot-system/}, and the experimental setup is shown in Fig. \ref{fig:video-data-teaser} and Fig. \ref{fig:simulation-data-teaser}.

To make this demonstration thorough, we also provide a brief (but comprehensive) overview of existing legibility frameworks, which is both a summary and an extension of previous reviews. 

%Additionally, we show that our framework is not just useful in an engineering context (build a robot that moves legibly), but that it can also help in a scientific context (why are some movements more legible than others). For this, we extract several factors (features), e.g., accumulated jerk, or total trajectory execution time, from existing literature which have previously been linked with legibility. Combining this list with commonly used confounds such as age or gender, we use the proposed evaluation method to test the importance of each factor in our two scenarios.

In sum, this paper makes the following contributions:

\begin{enumerate}
    \item The primary contribution is a new evaluation method that allows benchmarking of legibility frameworks (sec. \ref{sec:the-new-approach}).
    \item A secondary contribution is an overview of existing frameworks to generate legible motion (sec. \ref{sec:legibility-frameworks} and table \ref{tab:framework-overview}).
    \item Another secondary contribution is an evaluation of the quality of $10$ different legibility methods in two manipulation scenarios using the new evaluation method (sec. \ref{sec:video-data-results} and sec. \ref{sec:simulated-data-results}).
    \item A final secondary contribution is further experimental evidence showing that both the progress of a trajectory (sec. \ref{sec:video-data-results}) and the angle from which an observer views the trajectory (sec. \ref{sec:simulated-data-results}) influence a trajectories perceived legibility.
\end{enumerate}

We hope that this work can help with the identification of a more unified approach to modelling (and measuring) legibility. A unified approach will benefit researchers, on the one hand, in that it makes it easier to compare results, and it will benefit practitioners, on the other hand, in that it provides a tool to easily evaluate existing frameworks for their specific application scenario without having to rely on extensive experimentation.

\section{Related Work}\label{sec:related-work}
To get an overview of the existing work on legibility, we started with a keyword-based search on Scopus (see \textit{\nameref{appendix:scopus-search}} in the appendix). From this list we extracted relevant papers and - for each relevant paper - we checked other works citing it to maximize the amount of current, relevant work discovered. This yielded a total of $40$ relevant papers which we briefly summarize and categorize. We choose to keep this part of the paper short; however, there are several more detailed accounts of legibility and its related fields \cite{lichtenhaler2016, venture2019, anjomshoae2019, chakraborti2018b, wallkotter2020, reinhardt2019}.

On a high level, we found two big application domains of legibility: manipulation and navigation. Within the first, manipulation, research typically investigates pick-up or hand-over movements, i.e., how should a manipulator move to indicate that one (of several) objects is going to be picked up, or how should a manipulator move to indicate that it wishes to hand an object over to a user. Within the second, navigation, research typically investigates path-crossing movements, i.e., how should a mobile platform move to indicate that it yields priority to a human passerby that crosses the robot's path.

In the domain of manipulation, the focus of this paper, experimental work can broadly be divided into approaches that use an observer model \cite{sreedharan2017, sreedharan2019, dragan2013b, dragan2015, dragan2013a, chang2018, nikolaidis2016a, macnally2018, hough2017, kulkarni2019a, kulkarni2019b, zhang2017, zhou2017, mavrogiannis2018} and those that are model-free \cite{beetz2010, petruck2016, busch2017, knight2016, kuz2018, kebude2018, lamb2017a, lamb2018, lamb2017b, lamb2019}. 

A model-based approach scores trajectories using an explicit mathematical model of human expectations that is either cost-based (planning) or reward-based (reinforcement learning). Cost-based approaches solve an optimization problem to find the trajectory with minimum cost under some cost function (lower is better), whereas reward-based approaches solve a learning problem to find a policy that produces the trajectory with maximum reward under some reward function (higher is better). This difference matters in some applications; however, here we don't use the full optimization framework. Our approach only uses the cost/reward function, which can be used interchangeably (after eventual negation). As such, we review and compare approaches using both kinds of human models.

Model-free approaches, on the other hand, don't rely on explicit scoring and are generally more diverse. For example one model-free approach relies on directly modelling motion dynamics usable by low-level controllers \cite{lamb2017a, lamb2018, lamb2017b, lamb2019}, others use motion captured human data to extract (and imitate) their movement patterns \cite{kuz2018, beetz2010}.

Similarly, in the domain of navigation, we can find experimental work using a model-based approach. One example of a model-based approach is the HAMP framework \cite{sisbot2005, sisbot2007, kruse2012, kruse2014, lichtenthaler2012a, lichtenthaler2012b, lichtenthaler2014} that minimizes the sum of various HRI-based cost-terms to create a legible trajectory. While we do note that these approaches exist, we do not cover them further here, since the focus of this paper is on the manipulation domain.

Completing this summary, some authors also deviate from our definition of legibility and explore legibility through other modalities \cite{may2015, chadalavada2020, shreshta2016b, shrestha2016a}, e.g., lights, gaze, or scripted gestures with purely communicative intent (there is no/little progress on the task after the gesture finished executing).

\subsection{Legibility Frameworks}\label{sec:legibility-frameworks}
Focusing on papers discussing model-based approaches in manipulation, we identified $6$ papers that proposed frameworks to generate legible motion. 

For each of these papers, we extracted some high-level information (properties), which is shown in tab. \ref{tab:framework-overview}. In the table, the \textit{hyperparameters} column denotes if the framework can be fine-tuned via hyper-parameters. The \textit{planning space} column shows the space in which the optimization/learning takes place. \textit{Score} denotes if the framework uses a cost-based (lower is better) or reward-based (higher is better) metric to evaluate the score of a trajectory; this column strongly correlates with the associated framework using path-planning or reinforcement learning. \textit{Goal space} denotes if the framework creates a trajectory towards one (of many) discrete goals or towards a point in a continuous goal space. Finally, \textit{Eval Cost} is a (rough) estimate of the computational complexity involved in creating a trajectory through a framework.

\begin{table}[t]
\begin{adjustwidth}{-0.75in}{0in} % Comment out/remove adjustwidth environment if table fits in text column.
%\centering
\caption{Overview of existing Legibility Frameworks}
\label{tab:framework-overview}
\begin{tabular}{@{}lllllll@{}}
\toprule
Paper & Hyperparameters & Planning Space & Score & Goal Space & Eval Cost \\ \midrule
Bied legibility \cite{bied2020} & yes & discrete & reward & discrete & multiple \\
Bodden legibility \cite{bodden2018} & yes & multiple & cost & continous & normal \\
Busch legibility \cite{busch2017} & yes & joint & reward & discrete & very high \\
Dragan legibility \cite{dragan2013a, dragan2013b} & no & world & cost & discrete & high \\
Nikolaidis legibility \cite{nikolaidis2016a} & no & camera & cost & discrete & high \\
Zhao legibility \cite{zhao2020} & yes & world & reward & discrete & low \\
\midrule
Gulletta et al. \cite{gulletta2021} & yes & joint & cost & discrete & normal \\
Kulkarni et al. \cite{kulkarni2019a} & no & world & none & discrete & high \\
Lamb legibility \cite{lamb2019} & yes & joint & none & discrete & low \\
Sisbot et al. \cite{sisbot2007} & yes & world & cost & continous & normal \\
\bottomrule
\end{tabular}
\end{adjustwidth}
\end{table}

For $6$ of these papers we reimplemented the cost/reward function used by the framework in order to compare it using our novel method. Here, we noticed that some papers proposed more than one framework - typically by changing the used cost/reward function - resulting in a total of $10$ different frameworks to generate legible trajectories. A summary of these frameworks that, to the extent possible, unifies the notation is shown in table \ref{tab:legibility-equations}. In this table, $\gamma$, $\dot \gamma$, and $\dddot \gamma$ are the gripper's world space position, velocity, and jerk respectively, and $\gamma^P$ is the optimal trajectory under a cost function $C$. Further, $g_i$ is the $i$-th element in a set of goals $\mathbb{G}$, $g_0$ is (wlog) the intended goal, and $d^i(t) = |g_i(t) - \gamma(t)|$ is the distance to the $i$-th goal\footnote{Credit goes to \cite{zhao2020} for introducing this notation}. In \textit{Dragan Legibility}, \textit{Nikolaidis Legibility}, and \textit{Bied Legibility} $s$ is the robot's starting position, $T_{total}$ is the total time spent executing the trajectory, $\pi_\textrm{view}$ is the projection of a world space coordinate into the observers viewpoint (pixel space), and $P(g_0)$ is the user's prior belief that the trajectory will move to $g_0$. In \textit{Bodden Legibility} $\pi$ is any one of the three goal-projection functions introduced by \cite{bodden2018} (of which we implement and list two). $T_\textrm{world}^\textrm{eff}$ is the transformation from world space coordinates to end-effector space coordinates. In \textit{Zhao Legibility} $\delta(d_0 < \epsilon)$ is 1 if the distance to the intended goal is smaller than some threshold $\epsilon$ and 0 otherwise. Similarly, in \textit{Busch Legibility} $\delta(\mathrm{user\_correct?})$ is 1 if the user guessed the intended goal correctly, and 0 otherwise, and $T_\textrm{obs}$ is the time until the user made a guess. $\alpha$, $\beta$, $\sigma$, and $\epsilon$ are hyper-parameters, which we choose based on the values reported in the original papers.

\begin{table}[t]
\caption{Overview of Legibility Framework Equations}
\label{tab:legibility-equations}
\begin{tabular}{@{}ll@{}}
\toprule
Paper & Equation \\ \midrule
\multicolumn{2}{l}{Bodden Legibility \cite{bodden2018}} \\[.2em]
Trajectory Score & $\mathcal{L}[\gamma] = \int \alpha |g_0 - \pi(\gamma, \dot\gamma)|^2~\mathrm{d}t + \beta \int d_0^2~\mathrm{d}t + \epsilon \int |\dot \gamma|^2~\mathrm{d}t$ \\
Point & $\pi(\gamma, \dot\gamma) = \mathrm{d}^0$ \\
Velocity & $\pi(\gamma, \dot\gamma) = \angle(\dot\gamma, T_\textrm{world}^\textrm{eff}g_0)$ \\ \midrule
\multicolumn{2}{l}{Dragan Legibility \cite{dragan2013a, dragan2013b}} \\[.2em]
Trajectory Score & $\mathcal{L}_\textrm{Dragan}[\gamma] = \frac{\int P(g_0|\gamma_{s\to \gamma(t)})f~\mathrm{d}t}{\int f~\mathrm{d}t}$ \\
Predictability & $P(g_0|\gamma_{s\to \gamma(t)}) = \frac{\mathrm{exp}\left(C[\gamma^P_{s\to g_0}] - C[\gamma^P_{\gamma(t)\to g_0}]\right)}{\sum_{g\in\mathbb{G}} \mathrm{exp}\left(C[\gamma^P_{s\to g}] - C[\gamma^P_{\gamma(t)\to g}]\right)}P(g_0)$ \\
 & $C[\gamma] = \frac12 \int_t |\dot\gamma|^2$ \\ 
 & $f(t) = T_\textrm{total} - t$ \\\midrule
\multicolumn{2}{l}{Nikolaidis Legibility \cite{nikolaidis2016a}} \\[.2em]
Trajectory Score & $\mathcal{L}[\gamma] = \frac{\int P(\pi_\textrm{view}(g_0)|\pi_\textrm{view}(\gamma_{s\to \gamma(t)}))f~\mathrm{d}t}{\int f~\mathrm{d}t}$ \\ \midrule
\multicolumn{2}{l}{Busch Legibility \cite{busch2017}} \\[.2em]
Trajectory Score & $\mathcal{L}[\gamma] = T_\mathrm{obs} + \int 1~\mathrm{d}t + \beta~\delta(\mathrm{user\_correct?}) + \epsilon \int| \dddot \gamma |~\mathrm{d}t$ \\ \midrule
\multicolumn{2}{l}{Zhao Legibility \cite{zhao2020}} \\[.2em]
Trajectory Score & $\mathcal{L}[\gamma] = \int r_0 \delta(d_0 < \epsilon) + \beta~r_\textrm{legible}[\gamma]~\mathrm{d}t$ \\
fastApp & $r_\textrm{legible}[\gamma] = -d^0$ \\
effDist & $r_\textrm{legible}[\gamma] = e^{\frac{-t}{30}} \min_i\left(\log\left(\frac{|d_0 - d_i|}{|d_0 + 1|}+1\right)\mathrm{sgn}(d_0 - d_i)\right)$ \\ \midrule
\multicolumn{2}{l}{Bied Legibility \cite{bied2020}} \\[.2em]
Trajectory Score & $\mathcal{L}[\gamma] = \beta~r_\textrm{legible}[\gamma] + \epsilon\int |\dot\gamma|~\mathrm{d}t$ \\
Obs-L & $r_\textrm{legible}[\gamma] = \mathcal{L}_\textrm{Dragan}$ \\
Obs-P & $r_\textrm{legible}[\gamma] = \int P(g_0|\gamma_{s\to \gamma(t)})~\mathrm{d}t$ \\
Obs-D & $r_\textrm{legible}[\gamma] = \int \frac{\exp(-\sigma d_0)}{\sum_{g_i\in\mathbb{G}} \mathrm{exp}\left(-\sigma d_i\right)}~\mathrm{d}t$ \\ \bottomrule
\end{tabular}
\end{table}

\section{Materials and Methods}\label{sec:methods}
\subsection{The Traditional Approach}\label{sec:traditional-approach}
Another commonality of the identified framework papers is their evaluation method, which we call the traditional method in our work. To show the utility of the newly proposed framework, a paper (roughly) runs through the following steps:

\begin{enumerate}
    \item Generate a legible trajectory using the proposed method.
    \item Generate an alternative trajectory to the same goal using a second (baseline) framework, or expert knowledge.
    \item Use the generated trajectories in a user study and ask users to guess the desired goal.
    \item Run a statistical analysis (often t-test) to show that one trajectory is more legible than the other.
    \item Conclude that the framework producing the more legible trajectory (typically the proposed one) is superior.
\end{enumerate}

This is an appropriate approach to evaluate one specific framework; however, it does have two limitations:

The most severe limitation of this traditional approach is that it is very involved to compare multiple frameworks. Each comparison implies a new experiment, which requires new data collection and subsequent analysis. Here we investigate $10$ frameworks, which would would imply $45$ pairwise comparisons/experiments. This is expensive and time consuming. It can be somewhat mitigated by reusing previously collected data, if one (or both) frameworks have previously been tested in the same scenario; however, it still requires a new, dedicated data collection for each framework, and that is before considering any hyper-parameter tuning.

Another limitation is that the traditional approach typically only considers one movement between two (fixed) points; in other words, it determines the quality of the entire framework from a single trajectory. While studying such framework-optimal trajectory does have merit, it is also true that - in many situations - we use a motion planner precisely because we don't know the points between which the robot will move in advance. Just because a framework is better than another at moving between two specific points doesn't mean that this will be better in the entire workspace. However, the traditional approach does not allow an easy test of this.

To overcome the two limitations mentioned above we use a different, new evaluation method that we present in sec. \ref{sec:the-new-approach}. It specifically focuses on getting the most use out of the collected data and on minimizing the number of times that data has to be collected; i.e., it addresses the first limitation of the traditional approach. 
%The evaluation method can also overcome the second limitation, if the initial data collection (the first step of the method) covers the relevant workspace regions.

\begin{comment}
Our proposed way of comparing frameworks via a static set of trajectories addresses these limitations. Our approach decouples trajectories and frameworks, meaning that the human baseline only needs to be evaluated once, and then the performance of several frameworks can be evaluated without collecting additional data. We can investigate \textit{why} a framework performs poor, by performing an error analysis on trajectories where a framework and the human baseline mismatch. Further, we can directly evaluate the generality of a framework (in fact, many frameworks) by constructing the trajectory dataset from multiple environments.

The proposed approach is by no means a strict alternative to the traditional approach, but rather a generalization thereof. If we choose a dataset of two trajectories, and these trajectories happen to be the framework-optimal trajectories of two frameworks under study, then our approach reduces to the traditional approach for these two frameworks.
\end{comment}

\subsection{Human Baseline}\label{sec:human-baseline}
The foundation of the new approach is that it is possible to measure legibility in a data-driven manner: No matter which trajectory (or framework) we choose, it is always possible to evaluate a trajectory's legibility by asking users directly and by making them guess the intended goal. After asking multiple users, we can code their responses as correct/incorrect, compute the sample mean, and use that mean as an estimate of legibility. A mean close to 1 indicates a legible trajectory, and a mean close to 0 indicates an illegible trajectory. We call this the human-baseline, or gold-standard. It is expensive to compute - since it requires asking multiple users - but it is framework independent and accurate. As such, this is what we mean when we show \textit{legibility} in the result graphs in sec. \ref{sec:results}.

Indeed, a human-baseline is also computed in the traditional approach; however, this is only done for the framework-optimal trajectories of each framework being evaluated. One way that was previously used to obtain this human-baseline was to ask users to interrupt the robot's movement when they feel confident about the robot's target and to then ask them to announce the predicted target \cite{dragan2013a,kuz2018,busch2017}. This is, however, mostly done with physical robots - that come with a convenient emergency stop -, although it has been extended to videos of robots in online surveys.

In this work, we follow the same idea; however, instead of asking users to stop the movement, we only show a fraction of the trajectory, i.e., we artificially stop the movement before it reaches a goal and ask users to guess the goal at this stage. This approach, too, has been used previously to estimate the human-baseline of a trajectory \cite{nikolaidis2016a}.
In the first data collection, this allows us to assess how a user's belief changes during the execution of a trajectory by asking them to guess partial trajectories in increasing order of progression. Additionally, it has the advantage of being easier to implement in an online setting compared to the stop button approach.

What sets the new approach apart from the traditional one is that we do not limit evaluation of the human-baseline to framework-optimal trajectories. Instead, we first create a dataset of trajectories (independently of any framework), and then evaluate the human-baseline for all of these trajectories. Because the human-baseline for a trajectory is independent of the framework, this enables us to not have to collect any additional data from users after the first evaluation is complete, no matter the number of frameworks to be evaluated. We can reuse the same trajectories and human-baseline for every framework we evaluate.

\subsection{The New Approach}\label{sec:the-new-approach}
To prepare the evaluation of one (or many) frameworks using the new approach, we use the following steps \textit{once}:

\begin{enumerate}
    \item Generate a set of trajectories in the desired domain. There is no limitation on how the set of trajectories is chosen; however, they should cover relevant aspects, e.g., different viewing angles like in data collection 2 (sec. \ref{sec:simulated-data-collection}).
    \item Evaluate the Human-Baseline for each trajectory as described in sec. \ref{sec:human-baseline}.
\end{enumerate}

For the first step, we choose handcrafted trajectories here. This is motivated by our initial review of the literature. For each data collection, we choose trajectories based on experimental work to ensure that the created dataset covers previously identified important aspects of legible motion.
Following this up, we use the following steps to evaluate \textit{each framework}:

\begin{enumerate}[resume]
    \item Implement the scoring function of each framework.
    \item Evaluate the score of each trajectory chosen above using the implemented function.
    \item Compute the correlation between the framework's scores and the human-baseline. Here we use Spearman's Rank, but others (Pearson's R, Kendall's Tau) are also viable.
    \item Optional: Compute the correlation between the framework's scores and the scores of other frameworks.
\end{enumerate}

What is noteworthy in step 3 is that we omit a framework's generation mechanism and only evaluate the scoring function. The practical reason for this is that the generation mechanism will likely produce a trajectory that is not part of the initial dataset, and hence no human-baseline will exist for it. The theoretical reason for this is that the cost function is the framework's human model and encodes a framework's notion of legibility. Hence, testing it on a variety of trajectories and without the accompanying optimization scheme is a much more thorough evaluation.

Interestingly, we are not just limited to computing a notion of quality for a single framework (step 5), but can extend this to a notion of difference between frameworks (step 6). Instead of just computing a correlation with the human-baseline (quality), we can additionally compute a correlation with other frameworks. This correlation then indicates how alike two frameworks are in their measurement of legibility. It highlights the uniqueness of a framework, and can direct attention to aspects that might explain the advantage (or disadvantage) a framework. Hence, we can use the framework-framework correlation as an indicator to signal that we may wish to take a closer look at what this framework pair has in common and where they differ.

\subsection{Video Data Collection (Scenario 1)}\label{sec:video-data-collection}
As a first test of the new approach (sec. \ref{sec:the-new-approach}), and to get a sense of how much user feedback we need to gather per trajectory, we collect a dataset that focuses on varying trajectories as described below.

\subsubsection{Materials}\label{sec:data-collection-materials}
In this data collection we video record pick-up trajectories executed by a Franka Emika. The robot moves towards one of three goals (red/purple, green, blue), and is observed from the camera angle shown in Fig. \ref{fig:video-data-teaser}.

Here, we created $8$ different trajectories\footnote{Video of each motion: https://youtu.be/HyXl7djLJ5E} (see \nameref{appendix:video-data-trajectories} in the appendix). From each trajectory we create $3$ partials (up to $25\%$, $50\%$, and $75\%$ respectively), which we combine with the full trajectory (up to $100\%$) to obtain $4$ legibility evaluation points per trajectory. 

Concretely, we created a total of $5$ trajectories that move to the blue object, $2$ trajectories that move to the green object, and $1$ trajectory that moves to the red object. Our motivation for this is that - while accounting for cost and time constraints - we want to study different trajectories that move to the same goal, but also wish to have data for the same/similar movement to other goals so that we can compare the legibility of similarly shaped movements across different goals.

To choose the concrete path of each trajectory, we took inspiration from previous work: Motions \textit{Red 1}, \textit{Green 1}, and \textit{Blue 1} were inspired by the legibility described in \cite{bodden2018, zhao2016}. \textit{Green 2} was inspired by \cite{chang2018, kebude2018, petruck2016, zhao2016}. \textit{Blue2} was inspired by \cite{dragan2013a, bodden2018, zhao2016}. \textit{Blue 3} and \textit{Blue 4} were designed based on local discussions with colleagues. Finally, \textit{Blue 5} was designed to be intentionally confusing and illegible.

While most trajectories are inspired by existing legibility frameworks, we do not compute the trajectory based on any particular optimization scheme proposed by these works. Instead, the trajectories used here are handcrafted to be visually similar. While it is of course possible to use a set of framework-optimal trajectories, we explicitly avoid this to show that the new approach also works with trajectories that, according to a specific framework, are sub-optimal.

To compute the human baseline we video recorded each partial and asked users on AMT (Amazon Mechanical Turk) to guess the target object from the set of three objects. This was done using a multiple-choice question with one option per object (3 total) and the option "no answer". We compared this prediction with the actual target and coded/labeled the prediction as \textit{correct} or \textit{incorrect}. The legibility of a partial was then computed as the empirical mean of a participant guessing correctly.

\subsubsection{Design}
We used a mixed-design where a single user saw a single trajectory (between participants) but saw all partials of that trajectory (within participant). Each user, hence, watched $4$ videos, one for each level of progression for a single motion.

To ensure that we had sufficient data for our subsequent analyses, we estimated the number of participants needed using an a-priori power analysis. This was done for a reapeated measures ANOVA via the webpower package\footnote{https://cran.r-project.org/web/packages/WebPower/WebPower.pdf} in R, assuming a small effect size (0.25), the usual statistical power target (0.8). This produced a target sample size of $506$ participants. To this we added a margin of $26\%$ due to previously observed dropout from failed attention checks when using AMT, resulting in a target sample size of $640$ participants.

Ethical clearance was obtained from the counties regional ethics committee (Regionala etikprövningsnämnden i Uppsala, document no. 2018-503).

\subsubsection{Participants}
We recruited a total of $640$ participants ($\approx 80$ per trajectory) from AMT. To reduce the risk of participants misunderstanding the instructions, we only recruited participants residing in countries with English as the official language (US, Canada, UK, Australia, New Zealand). Further, we limited participation to workers that had previously completed at least $100$ assignments with $\geq 95\%$ approval rating. 

Participants ($M_{age}=36.43$, $SD=10.6$; $400$ male, $231$ female, $6$ non-binary, others unknown) received $0.6$ USD upon completion and took on average $2.6$ minutes to complete the survey. The equivalent median hourly wage was $16.33$ USD. Participants predominantly came from the US ($90\%$).

\subsubsection{Procedure}\label{sec:video-data-collection-procedure}
Participants first read an information sheet and consented to both participation and having their data processed. Next, they were randomly assigned to a condition corresponding to one of the $8$ trajectories. Participants then filled out a demographics questionnaire, and - after they completed it - were presented with the first video showing the 25\% partial of the assigned trajectory together with the legibility measure. This was repeated for the other partials in increasing order. Finally, participants were presented with a debriefing containing an explanation of the purpose of the data collection and a disclosure of the different conditions.

\subsection{Simulated Data Collection (Scenario 2)}\label{sec:simulated-data-collection}
As a second test of the new approach (sec. \ref{sec:the-new-approach}), we collected a dataset using videos generated using a roboics simulator. This was done due to the COVID-19 pandemic restricting access to the physical robot, and also because viewing angles can be controlled more accurately in a simulated environment. The aim of this data collection was to investigate the influence of viewing angle on legibility.

\subsubsection{Materials}
\begin{figure}
    \centering
    \includegraphics[width=\textwidth]{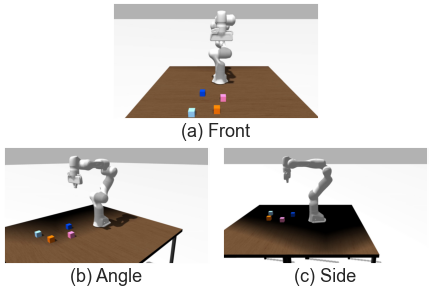}
    \caption{{\bf Viewing angles used in the simulation dataset.}}
    \label{fig:simulation-data-angles}
\end{figure}

In this data collection we used the Ignition simulator\footnote{https://ignitionrobotics.org/} to simulate pick-up trajectories executed by a Franka Emika. The robot moves towards one of four goals (light-blue, dark-blue, light-orange, pink), and is observed from one of three angles. The angles are shown in Fig. \ref{fig:simulation-data-angles}.

Similar to the previous data collection we created a set of fixed trajectories (see \nameref{appendix:simulation-data-trajectories} in the appendix). This time we used $2$ trajectories and from each created a single partial (up to $50\%$). This partial was, however, recorded from 3 different angles leading to a total of 6 different videos of partial trajectories.

To compute the human baseline, we generated a video for each partial and asked users on AMT to guess the target object from the set in the same was as we did for the video data collection.

\subsubsection{Design}
We used a mixed-design where a single user saw partials from a single angle (between participants) but saw all partials of each trajectory (within participant). Each user, hence, watched $2$ videos, one for each partial trajectory.

Similar to the video data collection (sec. \ref{sec:video-data-collection}) we computed the target number of participants a priori. However, we adjusted our estimate based on the new number of trajectory partials, our observation of (attention-check based) dropout in the previous data collection, and variance of ratings in the previously collected dataset. As such, we aimed for a total of $102$ participants.

Ethical clearance was, again, obtained from the counties regional ethics committee (Regionala etikprövningsnämnden i Uppsala, document no. 2018-503).

\subsubsection{Participants}
We recruited a total of $102$ participants ($\approx 34$ per angle) from AMT (Amazon Mechanical Turk) using the same criteria as in the video data collection (sec. \ref{sec:video-data-results}).

Participants ($M_{age}=35.56$, $SD=10.83$; $77$ male, $25$ female, $0$ other) received $0.8$ USD upon completion and took on average $3$ minutes to complete the survey. The equivalent median hourly wage was $20.77$ USD. Participants predominantly came from the US ($97.06\%$).

\subsubsection{Procedure}
The procedure, too, was similar to the video data collection (sec. \ref{sec:video-data-collection-procedure}). The only difference was that participants were assigned to different viewing angles (as opposed to trajectories), and saw all trajectory partials (as opposed to seeing all levels of progression).

\section{Results}\label{sec:results}
\subsection{Framework Benchmarks on Video Data (Scenario 1)}\label{sec:video-data-results}
\begin{figure}[t]
    %\centering
    \includegraphics[width=.98\textwidth]{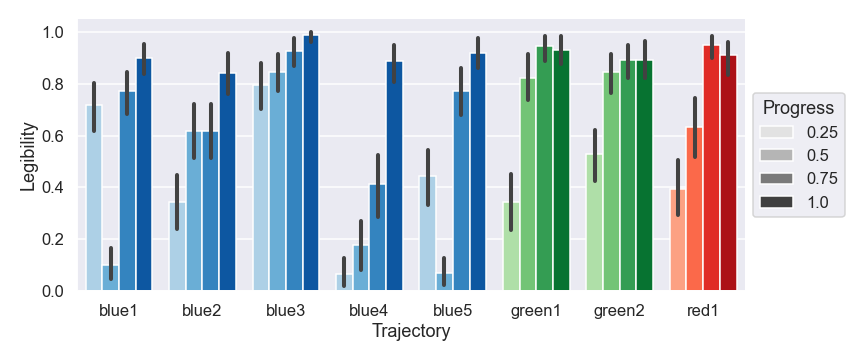}
    \caption{{\bf Human Baseline for the Video Dataset (Scenario 1)} Visualization of the rate at which humans correctly infer the correct goal for each partial trajectory (progress) of the video data collection. Legibility (y-axis) is measured from 0 to 1 and represents the odds of a human making the correct inference. Low values mark an illegible partial and high values mark a legible partial. The error bars show bootstrapped confidence intervals. Partials (x-axis) are grouped by the trajectory they were generated from and increasing saturation represents increasing progression (see legend). The color of a group indicates the true goal of the trajectory (blue, green, or red).}
    \label{fig:video-data-baseline-values}
\end{figure}

To analyze the collected data, we first computed the human baseline for each partial trajectory; this is shown in Fig. \ref{fig:video-data-baseline-values}. We can see that the legibility varies across partials and generally increases as more of the trajectory is revealed (increasing levels of progress). \textit{blue5}, the intentionally illegible trajectory, is indeed incorrectly inferred initially. The same is true for the trajectory \textit{blue4}, and \textit{blue1} takes an unexpected dip at the $.5$ progression level. Further, \textit{blue3} appears to be particularly legible.

\begin{figure}[t]
    \includegraphics[width=.98\textwidth]{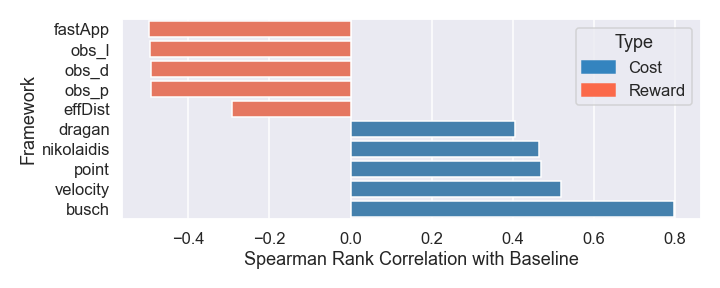}
    \caption{{\bf Framework benchmarks on the Video Dataset (Scenario 1)} Correlation between each framework's predicted legibility and the human baseline in the video data collection (scenario 1). The correlation (x-axis) is computed using Spearman rank on paired samples; one sample for each partial trajectory. Frameworks (y-axis) are color coded depending on type (see legend), i.e., depending on if they measure a cost (lower is better) or reward (higher is better).}
    \label{fig:video-data-framework-scores}
\end{figure}

Next, we computed the framework's scores for each partial trajectory and correlated it with the human baseline (Fig. \ref{fig:video-data-framework-scores}). For this, we use Spearman rank correlation, because frameworks scales differ in both range and distribution. As such, we prefer a non-parametric/rank-encoded measure of association like Spearman rank over the usual Pearson correlation coefficient\footnote{Kendall's $\tau$ would be an alternative and more conservative measure; however, we choose Spearman Rank here, since it is more widespread in the community.}.

Correlations hover around $.5$ across frameworks with \textit{Bush Legibility} scoring the largest magnitude and \textit{effDist Legibility} scoring the smallest magnitude. Unexpectedly so, all reward based frameworks show a negative correlation. This is \textit{not} because those frameworks use a reinforcement learning approach, but rather because of the choice of applied regularization. We discuss this in sec. \ref{sec:why-negative-correlation}. Overall, scores may have been higher had we further fine-tuned the frameworks on our specific dataset; however, we did not do this here, since we aim to compare out-of-the-box performance.

\begin{figure}[t]
    \centering
    \includegraphics[width=.45\textwidth]{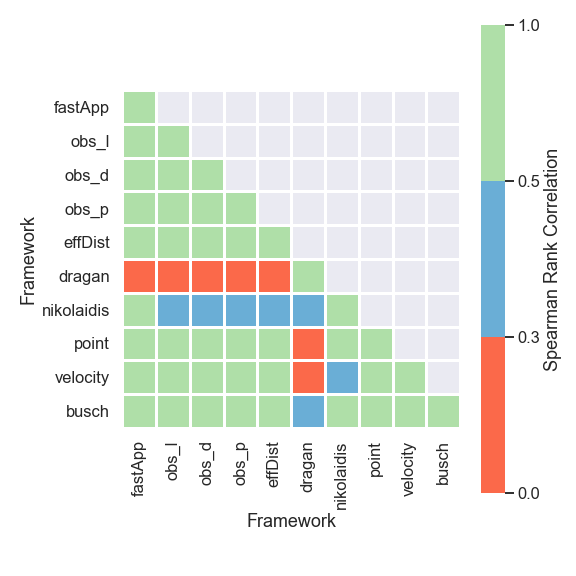}
    \caption{{\bf Framework correlation for the Video Dataset (Scenario 1)} Absolute correlation between each framework in the video data collection. The correlation (colorbar) is computed using Spearman rank on paired samples, one for each partial trajectory, and is binned into low/medium/high based on the usual cutoffs. High correlation indicates that two frameworks measure similar notions of legibility, and low correlation indicates that the frameworks measure different notions of legibility.}
    \label{fig:video-data-framework-correlation}
\end{figure}

On top of raw performance, we also investigated which frameworks rate the partials in a similar way. For this, we computed the correlation between each framework pair and binned it into low ($<.3$), medium $[.3, .5)$, and high $>.5$ correlation. Fig. \ref{fig:video-data-framework-correlation} shows the correlation for each such framework pair. We can see that most frameworks correlate highly, with exception of \textit{Dragan Legibility} and \textit{Nikolaidis Legibility}. \textit{Nikolaidis Legibility} shows large or medium correlation with the other frameworks. However, \textit{Dragan Legibility} - which many frameworks compare to - shows medium or low correlation with the other frameworks.

\subsection{Framework Benchmarks on Simulated Data (Scenario 2)}\label{sec:simulated-data-results}

\begin{figure}[t]
    \includegraphics[width=.95\textwidth]{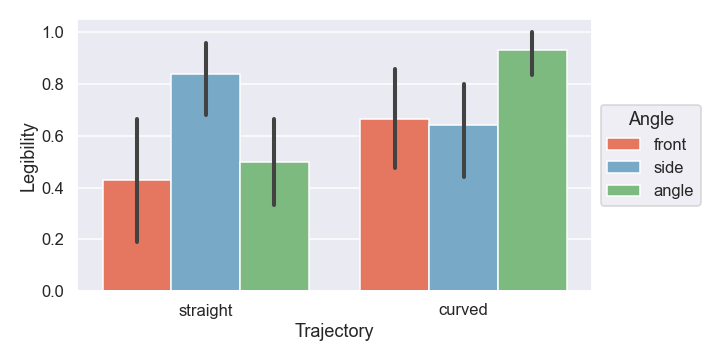}
    \caption{{\bf Human Baseline for the Simulation Dataset (Scenario 2)} Visualization of the rate at which humans correctly infer the correct goal when viewing each simulation dataset trajectory from different angles. Legibility (y-axis) is measured from 0 to 1 and represents the odds of a human making the correct inference. Low values indicate that the trajectory is illegible and high values indicate a legible trajectory. The error bars show bootstrapped confidence intervals. Different viewing angles (legend) are grouped by the trajectory (x-axis) that was used for better comparison.}
    \label{fig:simulation-data-values}
\end{figure}

\begin{figure}[t]
    \includegraphics[width=.95\textwidth]{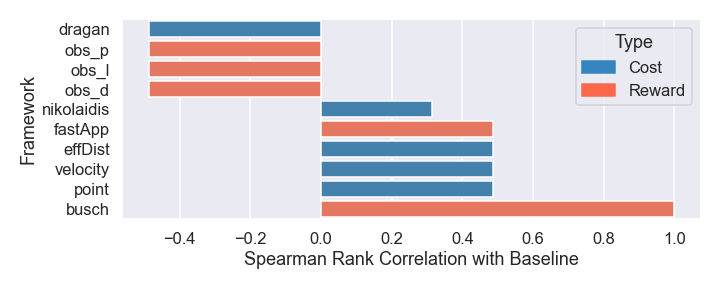}
    \caption{{\bf Framework benchmarks on the Simulation Dataset (Scenario 2)} Correlation between each framework's predicted legibility and the human baseline in the simulation dataset. The correlation (x-axis) is computed using Spearman rank on paired samples; one sample for each angle-trajectory pair. Frameworks (y-axis) are color coded depending on type (see legend), i.e., depending on if they measure a cost (lower is better) or reward (higher is better). For \textit{EffDist Legibility} no value is shown, as the correlation is zero (same framework score for each angle-trajectory pair).}
    \label{fig:simulation-data-scores}
\end{figure}

After computing the human baseline, we computed the trajectory ratings of each framework and correlated them with the previously computed human baseline. The result of this analysis is shown in Fig. \ref{fig:simulation-data-scores}. 

Although both the framework's score and the human-baseline are continuous variables, we can observe that the correlations appears as clusters. This is because none of the frameworks - with exception of \textit{Nikolaidis Legibility} and \textit{Busch Legibility} - account for the observer's angle, yet the angle does affect legibility ratings as we can see in Fig. \ref{fig:simulation-data-values}. This means that such a framework will score all viewing angles for the same trajectory with the same value, and that its correlation with the human baseline is thus only determined by the variance between trajectories. The dataset contains data from two different trajectories and, hence, we can observe correlation that appears clustered.

Further, we can observe that the value of \textit{Nikolaidis Legibility} is lower than that of the other frameworks with positive correlation. While the framework does accounts for angles (and, as such, doesn't cluster with the other frameworks) it is also based on \textit{Dragan Legibility}. \textit{Dragan Legibility} is negatively correlated for this dataset, which, combined with the fact that it doesn't account for angle, means that it (falsely) rates the straight trajectory as more legible one. This inaccuracy affects \textit{Nikolaidis Legibility}, too, and results in reduced performance compared to the other frameworks that correctly rate legibility.

\begin{figure}[t]
    \centering
    \includegraphics[width=.45\textwidth]{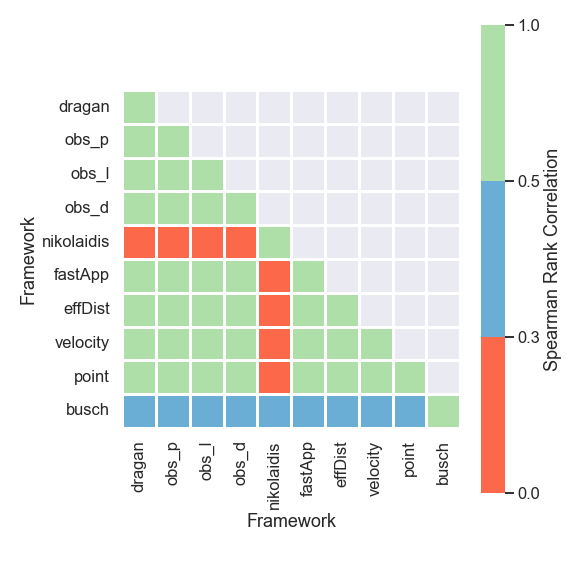}
    \caption{{\bf Framework correlation for the Simulation Dataset (Scenario 2).} Absolute correlation between each framework in the simulation dataset. The correlation (colorbar) is computed using Spearman rank on paired samples, one for each partial trajectory, and is binned into low/medium/high based on the usual cutoffs. High correlation indicates that two frameworks measure similar notions of legibility, and low correlation indicates that the frameworks measure different notions of legibility.}
    \label{fig:simulation-data-correlation}
\end{figure}

The final analysis of framework correlation (Fig. \ref{fig:simulation-data-correlation}) also confirms that \textit{Nikolaidis Legibility} and \textit{Busch Legibility} account for the angle whereas the other frameworks do not. The correlation between \textit{Nikolaidis Legibility} and every other framework is low, indicating that it measures a unique aspect of legibility. Considering that the unique feature of \textit{Nikolaidis Legibility} is precisely to account for the user's viewing angle, this result doesn't come as a surprise. Similarly, \textit{Busch Legibility} shows only a medium correlation with the other frameworks. This is because it uses human ratings to compute its score and, hence, indirectly accounts for viewing angle.

\section{Discussion}\label{sec:discussion}
\subsection{Framework Evaluation}\label{sec:benchmark-results}
One unexpected finding from our comparison between frameworks and the human baseline is that there is - overall - only medium correlation between frameworks and human ratings. One exception to this is \textit{Bush Legibility} \cite{busch2017} which relies on human ratings during evaluation, i.e., computes the human baseline in the processes of computing a framework score. As such, it will always result in an accurate result - as we can also see in our evaluation -. However, it is at least as expensive to compute as the human baseline, which is often undesirable, since we would hope to have a computational model for legibility that scales.

A potential explanation for this overall average agreement could be that we didn't fine-tune any hyperparameters. As shown in table \ref{tab:framework-overview}, most frameworks introduce various hyper-parameters to adjust a framework to a specific use-case, e.g., they use weights for different parts of a reward function. To push the performance of those frameworks, a future study could use a hyper-parameter search combined with a bootstrapping-based evaluation assuming there are enough computational resources available to compensate the increased computational demand of such an approach.

Another explanation could be that none of the frameworks fully model all important aspects of legibility. \textit{Busch Legibility}, which is computed from the human-baseline ratings, shows very high correlation, whereas the other frameworks - which don't require direct human feedback to compute - show lower performance. This would indicate that more experimental work is needed, to draw out the important factors of legibility that are currently not modeled by existing frameworks.

\subsection{Correlation between frameworks}
Looking at Fig. \ref{fig:video-data-framework-correlation} and Fig. \ref{fig:simulation-data-correlation} it is easy to see that not all frameworks agree with each other. This, however, does not mean that any particular framework is bad. In particular, just because \textit{Dragan Legibility} has low correlation with the other frameworks doesn't mean the framework doesn't measure legibility. Indeed, when looking at Fig. \ref{fig:video-data-framework-scores} and Fig. \ref{fig:simulation-data-scores}, we can see that \textit{Dragan Legibility} is on par with the other frameworks. 

Instead, low correlation indicates that the two frameworks measure different aspects of legibility. 

For Fig. \ref{fig:video-data-framework-correlation} this means that \textit{Dragan Legibility} measures a different/unique aspect of legibility. This could, however, be a false positive: Many frameworks use \textit{Dragan Legibility} as a baseline to compare themselves against. As such, these frameworks add features to differentiate themselves from and/or improve upon \textit{Dragan Legibility}. \textit{Nikolaidis Legibility} is an example of improving on \textit{Dragan Legibility}. It applies \textit{Dragan Legibility} in a different planning space (camera space vs the previous world space) resulting in better performance (higher correlation with the human baseline), but - consequentially - reduced correlation with \textit{Dragan Legibility}. \textit{Obs-l Legibility} is another example, but weights \textit{Dragan Legibility} quite low in comparison to a trajectory length based normalization term. Consequentially, we can only find a low correlation, as distance is only indirectly accounted for by \textit{Dragan Legibility}.

Similarily, for Fig. \ref{fig:simulation-data-correlation} we can see that \textit{Nikolaidis Legibility} and \textit{Busch Legibility} measure different aspects than the other frameworks. For \textit{Busch Legibility} this is easy to understand, as it is the only framework relying on direct human feedback. Looking at \textit{Nikolaidis Legibility}, it explicitly accounts for the observer's viewing angle, by projecting the trajectory into the observer's viewing plane while evaluating legibility. None of the other frameworks do this - despite viewing angle being important, as seen in Fig. \ref{fig:simulation-data-scores} - and hence we can observe low correlation between \textit{Nikolaidis Legibility} and the other frameworks. 

Summing up, most frameworks show high correlation with each other, indicating that they all measure legibility in a similar way. Exceptions to this are \textit{Dragan Legibility}, \textit{Nikolaidis Legibility}, and \textit{Busch Legibility}. For \textit{Busch legibility} this appears to be caused by the direct involvement of the human baseline, for \textit{Nikolaidis Legibility} this appears to be caused by the explicit accounting for angle, and for \textit{Dragan Legibility} this is most likely caused by it being the baseline for many of the other frameworks.

\subsection{The correlation negative of RL frameworks}\label{sec:why-negative-correlation}
A surprising finding was that all reward-based frameworks (see table \ref{tab:framework-overview}) show a negative correlation with the human baseline in Fig. \ref{fig:video-data-framework-scores}.

We can explain this by looking at how scores (rewards) are computed: They are, similar to some cost-based frameworks, computed as the sum of a legibility-related factor and a performance factor.  In our work, we implement this full (summed) reward function instead of only implementing the legibility related factor. 
We chose to do so, because we wanted to see how much legibility remains once the framework accounts for performance constraints while also optimizing for legibility. We used a similar approach in cost-based frameworks, where we implemented the cost function plus any regularization a framework applied.

The difference to cost-based frameworks is that the used performance factor is the negative trajectory length, i.e., the goal is to produce short, but legible trajectories. However, looking again at Fig. \ref{fig:video-data-baseline-values}, we can see that increasing progress increases the legibility score; at the same time, increased progress also means increased trajectory length. Hence, a factor/feature that is positively correlated with legibility is accounted for negatively. In the implemented frameworks the performance factor has a stronger influence than the legibility factor, and as such, a negative correlation with legibility is unsurprising.

At the same time, adding regularization terms based on objective performance factors, such as trajectory length, jerk, time, etc., undoubtedly helps. As such \textit{we might have to more thoroughly investigate what regularization to use}, because simple regularization schemes may unintentionally affect legibility as shown by our analysis.

\subsection{Framework Recommendation}\label{sec:how-to-choose}
In our experiments, the overall best performing framework was \textit{Bodden Legibility} \cite{bodden2018} using its \textit{velocity} metric. The framework is based around the novel idea of a goal space into which the robot's pose (given in joint space) is projected. The original paper suggests several (three) ways to compute this projection, of which we tested two. Both outperform most other frameworks on our video dataset and are on par for the simulation dataset, suggesting that there is utility in using an explicit goal space.

While \textit{Busch Legibility} \cite{busch2017} scores significantly higher than the other frameworks, we do not recommend using it in the general case due to its high evaluation cost. Computing the framework's score is done using explicit real-time human feedback, which is equivalent in complexity to computing the human baseline score. For interactive  scenarios, however, this approach is interesting, because it suggests updating the legibility model after each interaction. Unfortunately, our dataset does not include an interaction scenario, hence, evaluating framework performance for that domain remains a subject of future work.

\subsection{Limitations}\label{sec:limitations}
A first limitation of our approach is that the generality of our results hinges on the diversity of the trajectory dataset. We need to ensure that the dataset represents the domain we are interested in. This limitation is shared by any data-driven method; however, for trajectories specifically, it is still an open question what constitutes sufficient coverage.

A second limitation is that some frameworks were developed in a setting different from the one we apply it in. We aim to compare as vanilla a version of each framework as possible; however, changing context with minimal adaptation and fine-tuning may cause a framework to under-perform, i.e., if we would have tweaked it more, performance may have been better. That said, evaluating a tweaked approach is straight forward and - thanks to the new approach - directly comparable (we publish our trajectory data\footnote{https://github.com/FirefoxMetzger/2021-PLOS-ONE-Legibility}) to the performance numbers of the vanilla version that are reported here.

\section{Conclusion}\label{sec:conclusion}
In this paper we have proposed a new evaluation method for legibility frameworks and have applied it to various frameworks which we identified in a literature review. The method starts with generating trajectories independently of any framework, and evaluates them using user feedback, which we call the human baseline. It then evaluates each framework by comparing it to this human-baseline.

Applying the new approach, we found that most legibility frameworks score medium high, indicating that none of the frameworks fully account for all the factors that influence the perception of legible motion, and suggesting that more research is needed to discover and model them. Overall, we found that \textit{Bodden Legibility} \cite{bodden2018} (in particular when using the velocity metric) is the currently best performing framework.

We further discovered initial evidence that legibility frameworks based on reinforcement learning might currently take a sub-optimal approach to regularization. Reinforcement learning frameworks tend to penalize trajectory length (either explicitly, or via a cost-per-step mechanism); however, this may end up reducing their performance. We found that increased trajectory length correlates positively with progression towards the intended goal, which again correlates positively with increased legibility. As such, regularizing on trajectory length alone may decrease performance; however, explicit experimentation is required to confirm this and more research is needed to find better alternatives.

Finally, the data collections that we performed in the course of this work revealed two factors that influence legibility: Firstly, we found that a trajectories legibility changes as the trajectory progresses. Secondly the angle under which a trajectory is viewed, too, changes the trajectories legibility. Among the tested frameworks, only one (\textit{Nikolaidis Legibility}) explicitly accounted for the viewing angle. A fact that we could clearly see using the new approach to framework evaluation proposed here. Progression, too, has not been explicitly accounted for by current frameworks, although it has indirectly been part of their evaluation, e.g., by asking users to stop the robot once they are confident about the robot's intent. How to explicitly model the user's changing perception over the course of a trajectory, however, remains a subject for future work.

In sum, we have proposed a new approach to evaluating legibility frameworks, tested 
using several frameworks identified via literature review, and found that most frameworks perform satisfactory while also allowing for further improvement. We hope that this evaluation method can help in the creation of new, even better legibility frameworks, and in the identification of novel factors that influence legibility.

\section*{Supporting information}
\paragraph{S1}\label{appendix:scopus-search}
{\bf Scopus search strings.} For the initial keyword-based search we used two search strings. The first one targeted general work on legibility:

\begin{quote}
KEY ( "Human robot interaction"  OR  "Man machine systems"  OR  "Behavioral dynamics"  OR  "Human-robot collaboration"  OR  "behavioral dynamics"  OR  "human-robot collaboration"  OR  "Artificial intelligence"  OR  "Robot programming"  OR  "Classical planning"  OR  "Empirical evaluations"  OR  "Behavioral research"  OR  "Action selection"  OR  "Decision making"  OR  "Pick and place"  OR  "Communication"  OR  "Generalized models"  OR  "Off-line analysis"  OR  "Recognition models"  OR  "Human computer interaction"  OR  "coordination"  OR  "intent"  OR  "motion"  OR  "action interpretation"  OR  "formalism"  OR  "manipulation"  OR  "Trajectory optimization"  OR  "Motion planning"  OR  "motion planning"  OR  "trajectory optimization" ) AND
TITLE-ABS ( collaboration  OR  human  OR  observer  OR  collaborator  OR  plan  OR  justification  OR  "goal recognition" ) AND
TITLE-ABS ( dynamics  OR  dynamic  OR  motion  OR  action  OR  plan  OR  planner ) AND
TITLE-ABS ( model  OR  modelling  OR  system  OR  algorithm ) AND
TITLE-ABS ( transparent  OR  transparency  OR  express*  OR  explain*  OR  explanation  OR  legibility  OR  legible  OR  intent  OR  "goal recognition" ) AND
TITLE-ABS ( robot  OR  robotic  OR  agent ) AND
PUBYEAR > 2014 AND
LANGUAGE(english)
\end{quote}

The second one targeted work specifically on legibility in the domain of navigation, since the first search yielded predominantly work in a manipulation domain.

\begin{quote}
TITLE-ABS-KEY ( legible  OR  legibility  OR  intent )  AND  TITLE-ABS-KEY ( navigation  OR  navigational )  AND  TITLE-ABS-KEY ( robot*  OR  autonomous )
\end{quote}

\paragraph{S2}\label{appendix:video-data-trajectories}
{\bf Visualization of the trajectories used in the Video Data Collection.} 
\includegraphics[width=.8\textwidth]{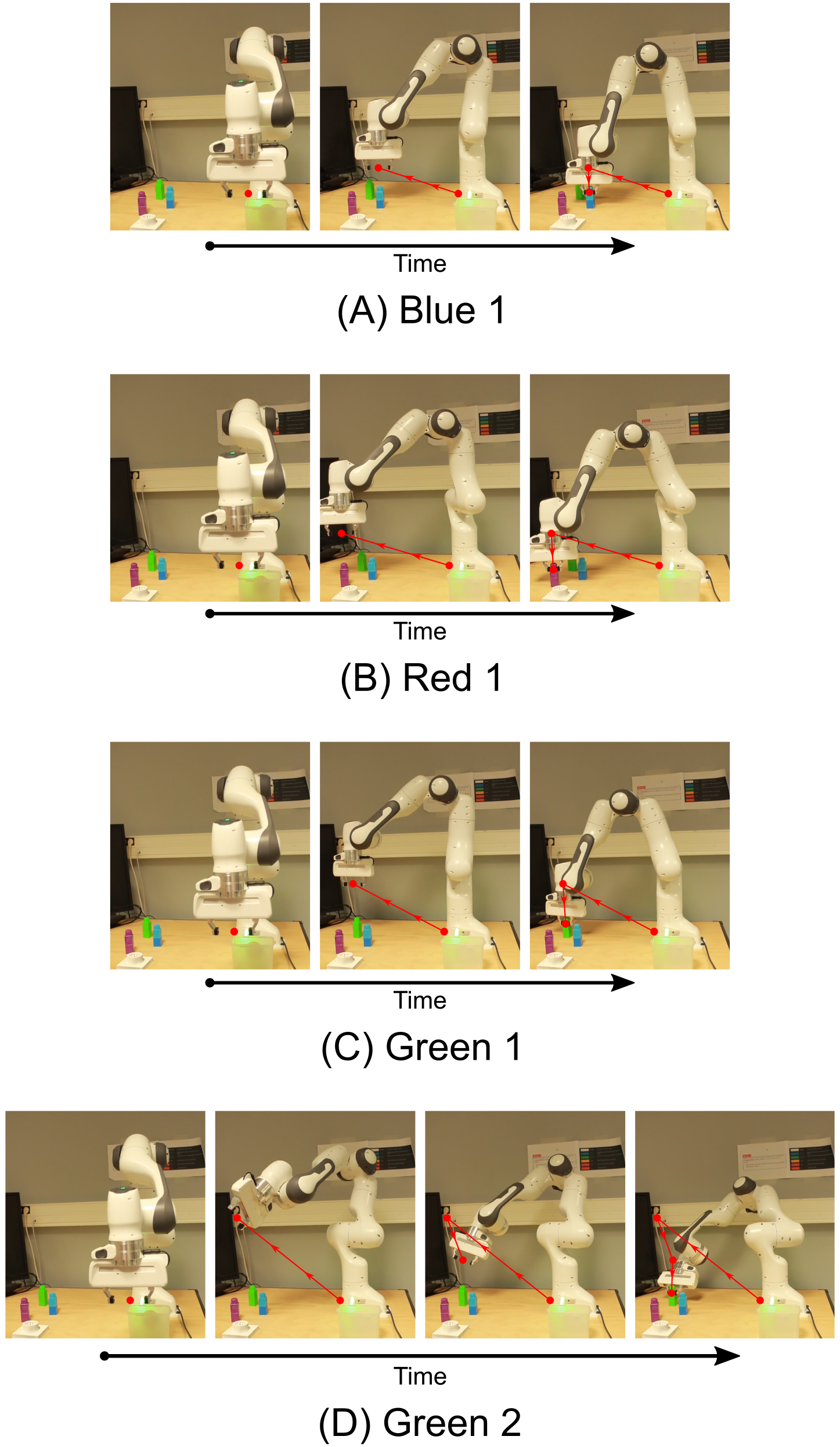}
\includegraphics[width=.8\textwidth]{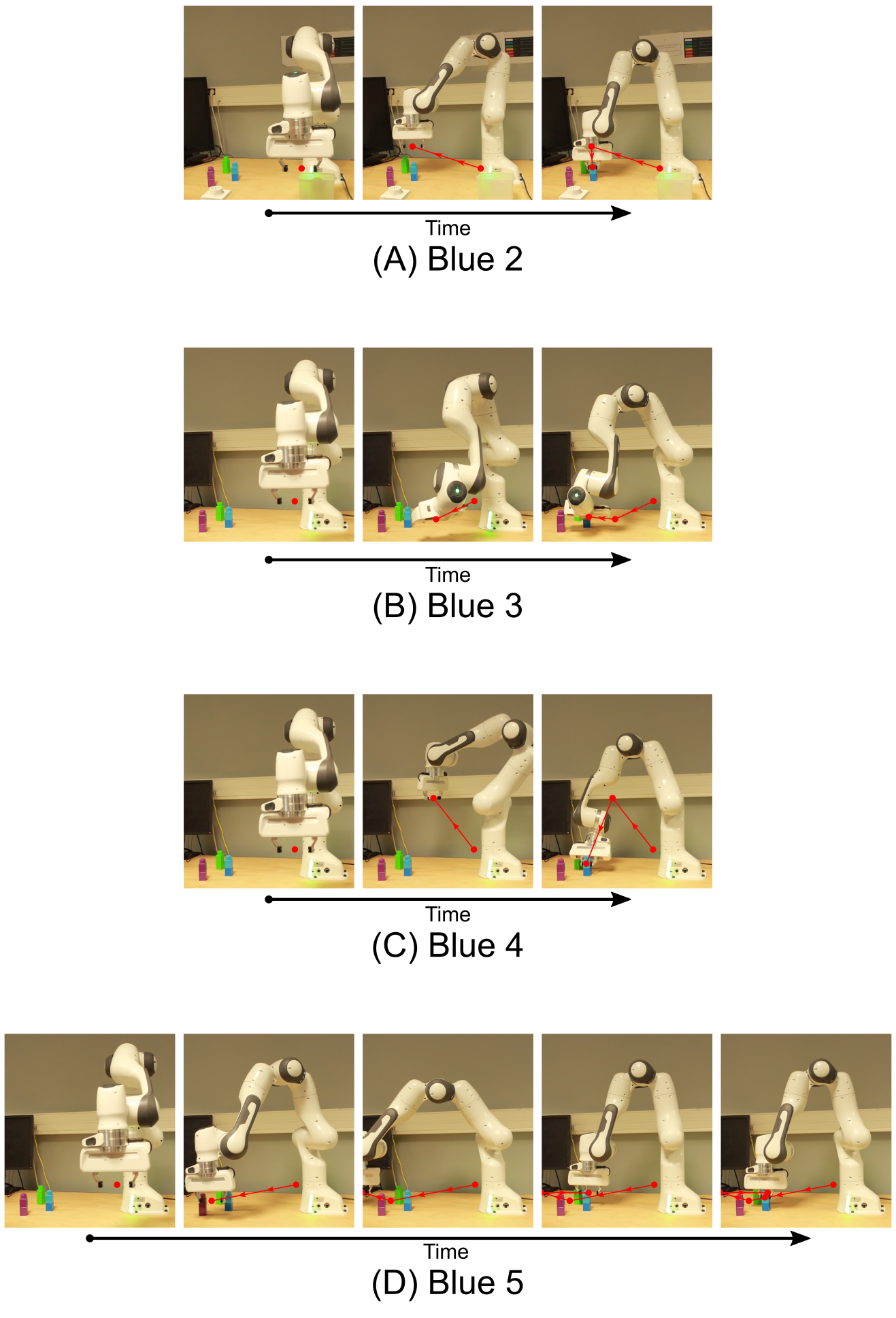}

\paragraph{S3}\label{appendix:simulation-data-trajectories}
{\bf Visualization of the trajectories used in the Simulated Data Collection.}
\includegraphics[width=\textwidth]{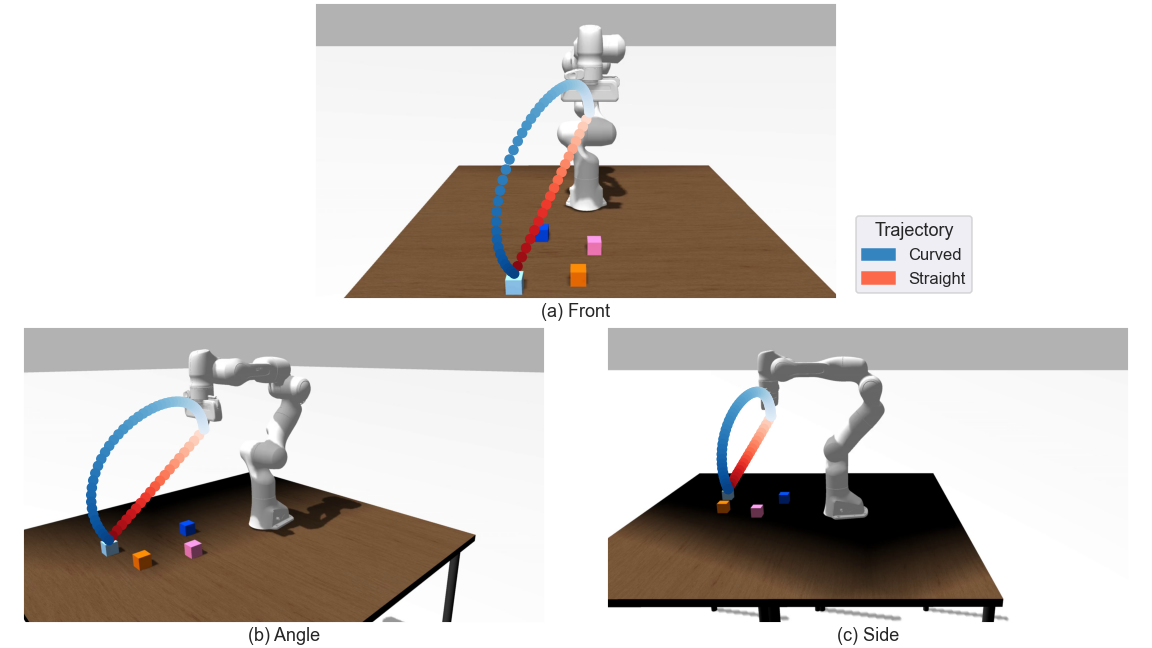}

\section*{Acknowledgments}
This project has received funding from the European Union's Horizon 2020 research and innovation programme under grant agreement No 765955.

% \nolinenumbers

% Compile your BiBTeX database using our plos2015.bst
% style file and paste the contents of your .bbl file
% here. See http://journals.plos.org/plosone/s/latex for 
% step-by-step instructions.
% 

% \bibliography{references}

\begin{thebibliography}{10}

\bibitem{wallkotter2020}
Wallk{\"{o}}tter S, Tulli S, Castellano G, Paiva A, Chetouani M.
\newblock {Explainable agents through social cues: A review}.
\newblock In: arXiv. arXiv; 2020. p.~23.

\bibitem{dautenhahn2005}
Dautenhahn K, Woods S, Kaouri C, Walters ML, Koay KL, Werry I.
\newblock {What is a robot companion - Friend, assistant or butler?}
\newblock In: 2005 IEEE/RSJ International Conference on Intelligent Robots and
  Systems, IROS; 2005. p. 1488--1493.

\bibitem{cha2018}
Cha E, Kim Y, Fong T, Mataric MJ, et~al.
\newblock A survey of nonverbal signaling methods for non-humanoid robots.
\newblock Foundations and Trends{\textregistered} in Robotics.
  2018;6(4):211--323.

\bibitem{lichtenhaler2016}
Lichtenth{\"{a}}ler C, Kirsch A.
\newblock {Legibility of Robot Behavior : A Literature Review}.
\newblock HAL-archives. 2016;1(1):1--22.

\bibitem{sreedharan2021}
Sreedharan S, Kulkarni A, Smith D, Kambhampati S.
\newblock A unifying bayesian formulation of measures of interpretability in
  human-AI interaction.
\newblock In: International Joint Conference on Artificial Intelligence; 2021.
  p. 4602--4610.

\bibitem{chakraborti2018b}
Chakraborti T, Kulkarni A, Sreedharan S, Smith DE, Kambhampati S.
\newblock {Explicability? Legibility? Predictability? Transparency? Privacy?
  Security? The Emerging Landscape of Interpretable Agent Behavior}.
\newblock In: arXiv: 1811.09722. unknown: ArXiv; 2018.

\bibitem{venture2019}
Venture G, Kuli{\'{c}} D.
\newblock {Robot Expressive Motions: A Survey of Generation and Evaluation
  Methods}.
\newblock ACM Transactions on Human-Robot Interaction. 2019;8(4):1--17.

\bibitem{anjomshoae2019}
Anjomshoae S, Najjar A, Calvaresi D, Fr{\"{a}}mling K.
\newblock {Explainable Agents and Robots: Results from a Systematic Literature
  Review}.
\newblock In: Proceedings of the 18th International Conference on Autonomous
  Agents and MultiAgent Systems (AAMAS '19). Aamas. Montreal: ACM; 2019. p.
  1078--1088.

\bibitem{reinhardt2019}
Reinhardt J, Schmidtler J, Bengler K.
\newblock {Corporate robot motion identity}.
\newblock In: Advances in Intelligent Systems and Computing. vol. 823. Springer
  Verlag; 2019. p. 152--164.

\bibitem{sreedharan2017}
Chakraborti T, Sreedharan S, Kambhampati S.
\newblock {Balancing explicability and explanations for human-aware planning}.
\newblock In: 28th International Joint Conference on Artificial Intelligence
  (IJCAI). vol. 2019-Augus. Macao: IJCAI; 2019. p. 1335--1343.
\newblock Available from: \url{http://arxiv.org/abs/1708.00543}.

\bibitem{sreedharan2019}
Sreedharan S, Chakraborti T, Muise C, Kambhampati S.
\newblock {Expectation-Aware Planning: A Unifying Framework for Synthesizing
  and Executing Self-Explaining Plans for Human-Aware Planning}.
\newblock yochan-labgithubio. 2019;.

\bibitem{dragan2013b}
Dragan AD, Lee KCT, Srinivasa SS.
\newblock {Legibility and Predictability of Robot Motion}.
\newblock In: 7th ACM/IEEE International Conference on Human-Robot Interaction
  (HRI). Tokyo: IEEE; 2013. p. 301--308.

\bibitem{dragan2015}
Dragan AD, Bauman S, Forlizzi J, Srinivasa SS.
\newblock {Effects of Robot Motion on Human-Robot Collaboration}.
\newblock In: ACM/IEEE International Conference on Human-Robot Interaction.
  vol. 2015-March; 2015. p. 51--58.

\bibitem{dragan2013a}
Dragan A, Srinivasa S.
\newblock {Generating Legible Motion}.
\newblock In: Robotics: Science and Systems IX. Berlin, Germany: none; 2013.
  p.~8.

\bibitem{chang2018}
{Lee Chang} M, Gutierrez RA, Khante P, {Schaertl Short} E, {Lockerd Thomaz} A.
\newblock {Effects of Integrated Intent Recognition and Communication on
  Human-Robot Collaboration}.
\newblock In: IEEE International Conference on Intelligent Robots and Systems;
  2018. p. 3381--3386.

\bibitem{nikolaidis2016a}
Nikolaidis S, Dragan A, Srinivasa S.
\newblock {Viewpoint-based legibility optimization}.
\newblock In: ACM/IEEE International Conference on Human-Robot Interaction.
  vol. 2016-April; 2016. p. 271--278.

\bibitem{macnally2018}
MacNally AM, Lipovetzky N, Ramirez M, Pearce AR.
\newblock {Action selection for transparent planning}.
\newblock In: Proceedings of the International Joint Conference on Autonomous
  Agents and Multiagent Systems, AAMAS. vol.~2; 2018. p. 1327--1335.

\bibitem{hough2017}
Hough J, Schlangen D.
\newblock {It's Not What You Do, It's How You Do It: Grounding Uncertainty for
  a Simple Robot}.
\newblock In: ACM/IEEE International Conference on Human-Robot Interaction.
  vol. Part F1271; 2017. p. 274--282.

\bibitem{kulkarni2019a}
Kulkarni A, Srivastava S, Kambhampati S.
\newblock {A Unified Framework for Planning in Adversarial and Cooperative
  Environments}.
\newblock Proceedings of the AAAI Conference on Artificial Intelligence.
  2019;33:2479--2487.
\newblock doi:{10.1609/aaai.v33i01.33012479}.

\bibitem{kulkarni2019b}
Kulkarni A, Srivastava S, Kambhampati S.
\newblock {Signaling Friends and Head-Faking Enemies Simultaneously: Balancing
  Goal Obfuscation and Goal Legibility}.
\newblock arXiv; 2019.

\bibitem{zhang2017}
Zhang YL, Sreedharan S, Kulkarni A, Chakraborti T, Zhuo HHH, Kambhampati S.
\newblock {Plan explicability and predictability for robot task planning}.
\newblock 2017 IEEE International Conference on Robotics and Automation (ICRA).
  2017; p. 1313--1320.
\newblock doi:{10.1109/ICRA.2017.7989155}.

\bibitem{zhou2017}
Zhou A, Hadfield-Menell D, Nagabandi A, Dragan AD.
\newblock {Expressive Robot Motion Timing}.
\newblock In: ACM/IEEE International Conference on Human-Robot Interaction.
  vol. Part F1271. New York, New York, USA: IEEE; 2017. p. 22--31.

\bibitem{mavrogiannis2018}
Mavrogiannis CI, Thomason WB, Knepper RA.
\newblock {Social Momentum: A Framework for Legible Navigation in Dynamic
  Multi-Agent Environments}.
\newblock In: ACM/IEEE International Conference on Human-Robot Interaction.
  IEEE Computer Society; 2018. p. 361--369.
\newblock Available from: \url{https://doi.org/10.1145/3171221.3171255}.

\bibitem{beetz2010}
Beetz M, Stulp F, Esden-Tempski P, Fedrizzi A, Klank U, Kresse I, et~al.
\newblock {Generality and legibility in mobile manipulation: Learning skills
  for routine tasks}.
\newblock Autonomous Robots. 2010;28(1):21--44.
\newblock doi:{10.1007/s10514-009-9152-9}.

\bibitem{petruck2016}
Petruck H, Kuz S, Mertens A, Schlick C.
\newblock {Increasing safety in human-robot collaboration by using
  anthropomorphic speed profiles of robot movements}.
\newblock In: Advances in Intelligent Systems and Computing. vol. 490. Springer
  Verlag; 2016. p. 135--145.

\bibitem{busch2017}
Busch B, Grizou J, Lopes M, Stulp F.
\newblock {Learning Legible Motion from Human–Robot Interactions}.
\newblock International Journal of Social Robotics. 2017;9(5):765--779.

\bibitem{knight2016}
Knight H, Simmons R.
\newblock {Laban head-motions convey robot state: A call for robot body
  language}.
\newblock In: Proceedings - IEEE International Conference on Robotics and
  Automation. vol. 2016-June; 2016. p. 2881--2888.

\bibitem{kuz2018}
Kuz S, Mertens A, Schlick CM.
\newblock {Anthropomorphic motion control of a gantry robot in assembly cells}.
\newblock Theoretical Issues in Ergonomics Science. 2018;19(6):738--751.

\bibitem{kebude2018}
Keb{\"{u}}de D, Eteke C, Sezgin TM, Akg{\"{u}}n B.
\newblock {Communicative cues for reach-to-grasp motions: From humans to
  robots: Robotics Track}.
\newblock In: Proceedings of the International Joint Conference on Autonomous
  Agents and Multiagent Systems, AAMAS. vol.~2. Stockholm, Sweden: ACM; 2018.
  p. 874--882.

\bibitem{lamb2017a}
Lamb M, Lorenz T, Harrison S, Kallen R, Minai A, Richardson M.
\newblock {Behavioral Dynamics and Action Selection in a Joint Action
  Pick-and-Place Task}.
\newblock In: Proceedings of the Annual Meeting of the Cognitive Science
  Society (CogSci). vol.~1. CogSci; 2017. p. 2506--2511.

\bibitem{lamb2018}
Lamb M, Mayr R, Lorenz T, Minai AA, Richardson MJ.
\newblock {The Paths We Pick Together: A Behavioral Dynamics Algorithm for an
  HRI Pick-and-Place Task}.
\newblock In: Proceedings of the 2018 ACM/IEEE International Conference on
  Human-Robot Interaction. Chicago: IEEE; 2018. p. 165--166.

\bibitem{lamb2017b}
Lamb M, Kallen RW, Harrison SJ, {Di Bernardo} M, Minai A, Richardson MJ.
\newblock {To pass or not to pass: Modeling the movement and affordance
  dynamics of a pick and place task}.
\newblock Frontiers in Psychology. 2017;8(JUN).
\newblock doi:{10.3389/fpsyg.2017.01061}.

\bibitem{lamb2019}
Lamb M, Nalepka P, Kallen RW, Lorenz T, Harrison SJ, Minai AA, et~al.
\newblock {A Hierarchical Behavioral Dynamic Approach for Naturally Adaptive
  Human-Agent Pick-and-Place Interactions}.
\newblock Complexity. 2019;2019.

\bibitem{sisbot2005}
Sisbot EA, Alami R, Simeon T, Dautenhahn K, Walters M, Woods S, et~al.
\newblock {Navigation in the presence of humans}.
\newblock In: Proceedings of 2005 5th IEEE-RAS International Conference on
  Humanoid Robots. vol. 2005; 2005. p. 181--188.

\bibitem{sisbot2007}
Sisbot EA, Marin-Urias KF, Alami R, Sim{\'{e}}on T.
\newblock {A human aware mobile robot motion planner}.
\newblock In: IEEE Transactions on Robotics. vol.~23; 2007. p. 874--883.

\bibitem{kruse2012}
Kruse T, Basili P, Glasauer S, Kirsch A.
\newblock {Legible robot navigation in the proximity of moving humans}.
\newblock In: Proceedings of IEEE Workshop on Advanced Robotics and its Social
  Impacts, ARSO; 2012. p. 83--88.

\bibitem{kruse2014}
Kruse T, Kirsch A, Khambhaita H, Alami R.
\newblock {Evaluating directional cost models in navigation}.
\newblock In: ACM/IEEE International Conference on Human-Robot Interaction.
  IEEE Computer Society; 2014. p. 350--357.

\bibitem{lichtenthaler2012a}
Lichtenthaler C, Lorenzy T, Kirsch A.
\newblock {Influence of legibility on perceived safety in a virtual human-robot
  path crossing task}.
\newblock In: Proceedings - IEEE International Workshop on Robot and Human
  Interactive Communication; 2012. p. 676--681.

\bibitem{lichtenthaler2012b}
Lichtenth{\"{a}}ler C, Lorenz T, Karg M, Kirsch A.
\newblock {Increasing perceived value between human and robots - Measuring
  legibility in human aware navigation}.
\newblock In: Proceedings of IEEE Workshop on Advanced Robotics and its Social
  Impacts, ARSO; 2012. p. 89--94.

\bibitem{lichtenthaler2014}
Lichtenth{\"{a}}ler C, Kirsch A.
\newblock {Goal-predictability vs. trajectory-predictability - Which legibility
  factor counts}.
\newblock In: ACM/IEEE International Conference on Human-Robot Interaction.
  IEEE Computer Society; 2014. p. 228--229.

\bibitem{may2015}
May AD, Dondrup C, Hanheide M.
\newblock {Show me your moves! Conveying navigation intention of a mobile robot
  to humans}.
\newblock In: 2015 European Conference on Mobile Robots, ECMR 2015 -
  Proceedings; 2015.

\bibitem{chadalavada2020}
Chadalavada RT, Andreasson H, Schindler M, Palm R, Lilienthal AJ.
\newblock {Bi-directional navigation intent communication using spatial
  augmented reality and eye-tracking glasses for improved safety in
  human–robot interaction}.
\newblock Robotics and Computer-Integrated Manufacturing. 2020;61.

\bibitem{shreshta2016b}
Shrestha MC, Kobayashi A, Onishi T, Yanagawa H, Yokoyama Y, Uno E, et~al.
\newblock {Exploring the use of light and display indicators for communicating
  directional intent}.
\newblock In: IEEE/ASME International Conference on Advanced Intelligent
  Mechatronics, AIM. vol. 2016-Septe; 2016. p. 1651--1656.

\bibitem{shrestha2016a}
Shrestha MC, Kobayashi A, Onishi T, Uno E, Yanagawa H, Yokoyama Y, et~al.
\newblock {Intent communication in navigation through the use of light and
  screen indicators}.
\newblock In: ACM/IEEE International Conference on Human-Robot Interaction.
  vol. 2016-April; 2016. p. 523--524.

\bibitem{bied2020}
Bied M, Chetouani M.
\newblock {Exploring the difference between solving and teaching in
  sensorimotor tasks}.
\newblock In: ACM/IEEE International Conference on Human-Robot Interaction;
  2020. p. 139--141.

\bibitem{bodden2018}
Bodden C, Rakita D, Mutlu B, Gleicher M.
\newblock {A flexible optimization-based method for synthesizing
  intent-expressive robot arm motion}.
\newblock International Journal of Robotics Research. 2018;37(11):1376--1394.
\newblock doi:{10.1177/0278364918792295}.

\bibitem{zhao2020}
Zhao X, Fan T, Wang D, Hu Z, Han T, Pan J.
\newblock {An Actor-Critic Approach for Legible Robot Motion Planner}.
\newblock In: Proceedings - IEEE International Conference on Robotics and
  Automation; 2020.

\bibitem{gulletta2021}
Gulletta G, Silva ECe, Erlhagen W, Meulenbroek R, Costa MFP, Bicho E.
\newblock {A Human-like Upper-limb Motion Planner: Generating naturalistic
  movements for humanoid robots}.
\newblock International Journal of Advanced Robotic Systems. 2021;18(2).

\bibitem{zhao2016}
Zhao M, Shome R, Yochelson I, Bekris K, Kowler E.
\newblock {An experimental study for identifying features of legible
  manipulator paths}.
\newblock In: Springer Tracts in Advanced Robotics. vol. 109; 2016.

\end{thebibliography}

\end{document}